%% file: corpipe25.tex
\pdfoutput=1
\documentclass[11pt]{article}

\usepackage[preprint]{acl}

\usepackage{times}
\usepackage{latexsym}

\usepackage[T1]{fontenc}
\usepackage[utf8]{inputenc}

\usepackage{microtype}

\usepackage{inconsolata}

\usepackage{graphicx}

\usepackage{amssymb}
\usepackage{booktabs}
\usepackage{makecell}
\usepackage{multirow}
\usepackage{xcolor}
\usepackage{xspace}

\newcommand{\CRAC}{CRAC 2025 Shared Task\xspace}
\newcommand{\CRAClong}{CRAC 2025 Shared Task on Multilingual Coreference Resolution\xspace}
\newenvironment{citemize}{\begin{list}{$\bullet$}{\topsep=.1\smallskipamount\itemsep=0pt\parsep=1pt\labelwidth=.5em}}{\end{list}}

\title{CorPipe at CRAC 2025: Evaluating Multilingual Encoders\\for Multilingual Coreference Resolution}

\author{Milan Straka \\
  Charles University, Faculty of Mathematics and Physics \\
  Institute of Formal and Applied Linguistics \\
  Malostranské nám. 25, Prague, Czech Republic \\
  \texttt{straka@ufal.mff.cuni.cz}
  }

\usepackage{fancyhdr}
\fancypagestyle{officialbibref}{\fancyhf{}\fancyheadoffset[L,R]{50pt}\fancyhead[C]{This paper was published at \textbf{CODI-CRAC 2025} -- please cite the published version {\small\url{https://aclanthology.org/2025.crac-1.11}}.}\fancyfoot[C]{\normalsize\thepage}\addtolength{\topmargin}{-30pt}\addtolength{\headsep}{30pt}}
\begin{document}
\thispagestyle{officialbibref}
\maketitle
\begin{abstract}
  We present CorPipe 25, the winning entry to the \CRAClong. This fourth
  iteration of the shared task introduces a new LLM track alongside the
  original unconstrained track, features reduced development and test sets to
  lower computational requirements, and includes additional datasets. CorPipe
  25 represents a complete reimplementation of our previous systems, migrating
  from TensorFlow to PyTorch. Our system significantly outperforms all other
  submissions in both the LLM and unconstrained tracks by a substantial margin
  of 8 percentage points. The source code and trained models are publicly
  available at {\small\url{https://github.com/ufal/crac2025-corpipe}}.
\end{abstract}

\section{Introduction}

Coreference resolution seeks to identify and cluster multiple references to the
same entity within text. The \CRAClong~\citep{novak-etal-2025-findings}
represents the fourth iteration of this shared task, designed to advance
research in multilingual coreference resolution across diverse languages and
domains. Building upon the CorefUD 1.3 collection, this year's task introduces
several notable changes: a new LLM track that relies on large language models
(LLMs) for coreference resolution, reduced development and test sets (minidev
and minitest) to lower computational demands, and the inclusion of additional
datasets expanding language coverage.

As in the previous year, the submitted systems must also predict the
\textit{empty nodes}, which represent elided elements that are not explicitly
present in the surface text but are necessary for coreference analysis.
Empty nodes are especially important in pro-drop languages (like Slavic and
Romance languages), where pronouns can be dropped from a sentence when
they can be inferred, for example according to verb morphology, as in the \hbox{Czech example}
\textit{``Řekl,~že~nepřijde''}, translated as \textit{``(He) said that (he)
won't come''}.

CorPipe 25, our submission to the \CRAC, represents a complete reimplementation
of our previous winning
systems~\citep{straka-2024-corpipe,straka-2023-ufal,straka-strakova-2022-ufal},
transitioning from TensorFlow to PyTorch while preserving the architecture
that has proven successful. Our system employs a three-stage pipeline
approach: first predicting empty nodes,\footnote{Our empty node prediction
system was provided to all participants as a baseline implementation.} then
detecting mentions, and finally performing coreference linking through
antecedent maximization on the identified spans. As in previous CorPipe
versions, mention detection and coreference linking are trained jointly
using a shared pretrained encoder model, and all models are fully
multilingual, trained across all available corpora.

Our contributions are as follows:
\begin{citemize}
  \item We present the winning entry to the \CRAC, surpassing other
    participants in both tracks by a substantial margin of 8 percentage points.
  \item We provide a complete reimplementation of CorPipe in PyTorch. The
    reimplementation enables us to leverage more pretrained multilingual
    models, allowing us to perform an evaluation of various models and
    providing insights into their relative performance for coreference
    resolution across diverse languages.
  \item We present performance comparisons between TensorFlow and PyTorch
    implementations, demonstrating the practical benefits of the migration.
  \item The CorPipe 25 source code is released at
    {\small\url{https://github.com/ufal/crac2025-corpipe}} under an
    open-source license. Three pretrained multilingual models of different sizes
    are also released, under the CC BY-NC-SA licence.
\end{citemize}

\section{Related Work}

\textbf{Neural Coreference Resolution}~~
Neural coreference resolution has been dominated by span-based approaches since
the seminal work of \citet{lee-etal-2017-end}, who introduced an end-to-end
neural model that jointly performs mention detection and coreference
resolution. This approach was further refined by \citet{lee-etal-2018-higher}
with coarse-to-fine inference, significantly improving both efficiency and
accuracy. \citet{joshi-etal-2020-spanbert} demonstrated substantial
improvements by incorporating SpanBERT \citep{joshi-etal-2019-bert},
a pretrained model specifically designed for span prediction tasks.

Alternative paradigms have emerged to address the limitations of span-based
methods. \citet{wu-etal-2020-corefqa} formulated coreference as
a question-answering task, while \citet{liu-etal-2022-autoregressive} introduced
a specialized autoregressive system and \citet{bohnet-etal-2023-coreference}
employed a text-to-text paradigm. However, all these architectures
must evaluate the trained model repeatedly during processing of a single
sentence.

\textbf{Word-Level Coreference Resolution}~~
A significant departure from span-based approaches came with
\citet{dobrovolskii-2021-word}, who proposed word-level coreference resolution,
which represents mentions by their head-words only. The approach has been
extended by \citet{doosterlinck-etal-2023-caw} with CAW-coref, which introduces
conjunction-aware handling to better manage complex mention structures. More
recently, \citet{liu-etal-2024-mscaw} proposed MSCAW-coref that aims to work in
a multilingual setting and accounts for singleton mentions.
This approach has been adopted by Stanza \citep{qi-etal-2020-stanza},
a widely-used Python natural language processing toolkit.

\textbf{Multilingual Coreference Resolution}~~
The CRAC shared tasks on multilingual coreference resolution
\citep{zabokrtsky-etal-2022-findings,zabokrtsky-etal-2023-findings,novak-etal-2024-findings,novak-etal-2025-findings}
have been instrumental in advancing the field, providing standardized
evaluation framework, the CorefUD dataset \citep{CorefUD1.3}, and
a multilingual baseline \citep{prazak-etal-2021-multilingual}.

Previous versions of CorPipe have participated in all CRAC shared
tasks, evolving from basic multilingual models
\citep{straka-strakova-2022-ufal} to incorporating larger contexts
\citep{straka-2023-ufal} and performing zero mention prediction from raw text
\citep{straka-2024-corpipe}.

\section{Architecture}

Our system is essentially a PyTorch reimplementation of CorPipe
24~\citep{straka-2024-corpipe}.

\begin{figure}[t]
    \centering
    \includegraphics[width=\hsize]{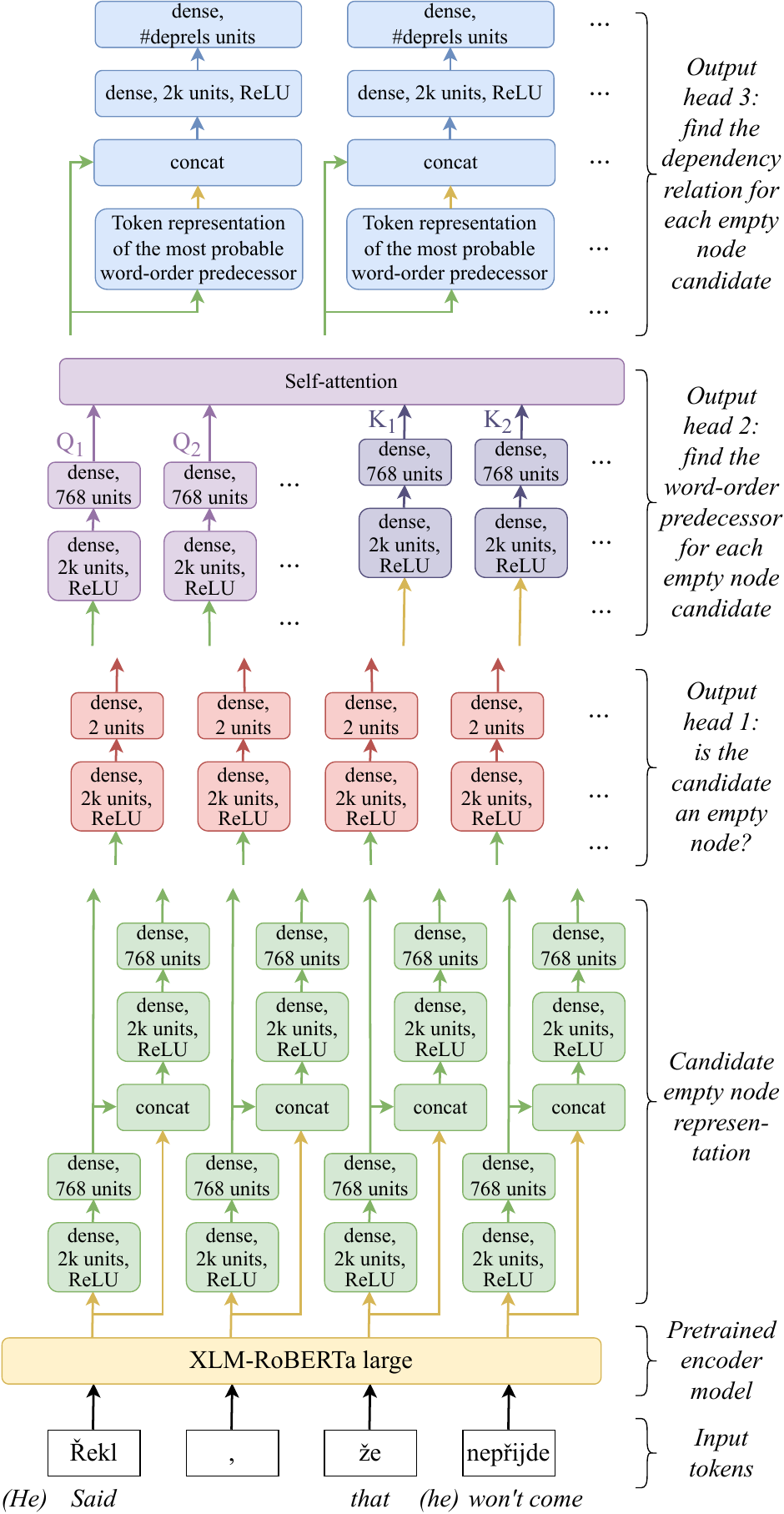}
    \caption{The system architecture of the empty node prediction baseline.
    Every ReLU activation is followed by a dropout layer with a dropout rate of 50\%.}
    \label{fig:empty_nodes_baseline_architecture}
\end{figure}

\begin{figure*}[t]
    \centering
    \includegraphics[width=.86\hsize]{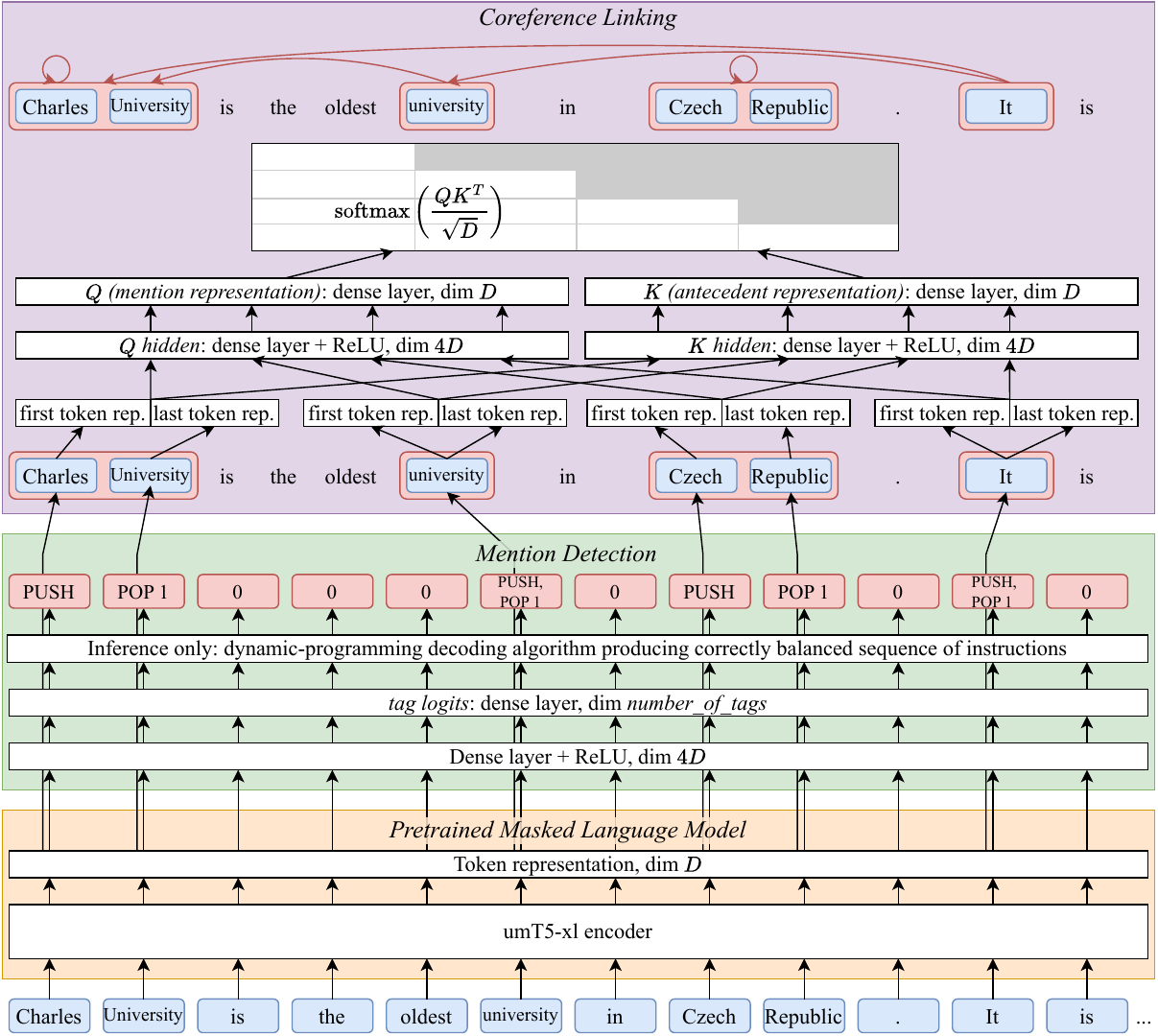}
    \caption{The CorPipe 25 model architecture.}
    \label{fig:corpipe25_architecture}
\end{figure*}

\textbf{Empty Nodes Baseline}~~
First, empty nodes are predicted using a baseline system that was available
to all shared task participants. The architecture of this system is illustrated
in Figure~\ref{fig:empty_nodes_baseline_architecture}.

Our approach for empty node prediction focuses on generating the essential
information required for coreference evaluation: the word order position
(determined by which input word the empty node follows), along with the
dependency head and dependency relation. We do not predict forms or lemmas,
even when available in training data. The model operates non-autoregressively,
predicting up to two empty nodes per input word, with each input word serving
as the potential dependency head.

\looseness-1
The architecture processes tokenized input through
a XLM-RoBERTa-large~\citep{conneau-etal-2020-unsupervised}, representing each
word by its first subword embedding. For each word, we generate two empty node
candidates: the first through a dense-ReLU-dropout-dense module
(768$\rightarrow$2k$\rightarrow$768 units), and the second by concatenating the
first candidate with the input word representation and applying an analogous
transformation. The candidates are processed by three heads, each following its own 2k-unit
ReLU layer and dropout: (1) binary classification for empty node existence, (2)
self-attention for word order position selection, and (3) dependency relation
classification using the candidate representation concatenated with the
embedding of the most likely word preceding it.

Training employs a single multilingual model with Adam
optimizer~\citep{kingma-and-ba-2015-adam} for 20 epochs of 5\,000 batches
(64 sentences each). The learning rate linearly increases to 1e-5 in the first
epoch and then decays to zero in the rest of the training following cosine
decay~\citep{loshchilov-etal-2017-sgdr}. Sentences
are sampled from all empty node corpora, proportionally to the square root of
corpus size. Training required 19 hours on a single L40 GPU with 48GB RAM.

\looseness-1
The source code is released under the MPL license at
{\small\url{https://github.com/ufal/crac2025_empty_nodes_baseline}}, together
with the full set of hyperparameters used. The trained model
is available under the CC BY-SA-NC license at
{\small\url{https://www.kaggle.com/models/ufal-mff/crac2025_empty_nodes_baseline/}}.
Finally, the minidev and minitest sets of the \CRAC with predicted
empty nodes are available to all participants.

\looseness-1
\textbf{Coreference Resolution}~~
Once the empty nodes have been predicted, we employ coreference resolution
system based on CorPipe 23 from~\citet{straka-2023-ufal}. The architectural
overview is shown in Figure~\ref{fig:corpipe25_architecture} and summarized
below; detailed implementation specifics are available in the referenced work.

Our model processes documents sentence-by-sentence. To maximize available
context for each sentence, we expand it with preceding tokens and at most 50 subsequent
tokens, constrained by the maximum segment length (512 or 2\,560 tokens). Input tokens
first pass through a pretrained multilingual encoder. Subsequently, we predict
coreference mentions using an enhanced BIO encoding scheme that handles
potentially overlapping span sets. Each identified mention is then encoded as
a concatenation of its boundary tokens (first and last), and coreference
links are established through a self-attention mechanism that determines the
most probable antecedent for each mention (including self-reference utilized by
first entity mentions).

We employ different segment sizes during training versus inference: training
always uses \hbox{512-token} segments, while inference leverages extended
2\,560-token segments (with the exception of two PROIEL corpora
always using 512 tokens), exploiting relative positional
encoding capabilities for improved long-range context modeling.

\begin{table}[t]
  \setlength{\tabcolsep}{4.5pt}
  \begin{tabular*}{\hsize}{lrccc}
    \toprule
      Model & Params & \makecell[c]{Batch\\Size} & \makecell[c]{Learning\\Rate} & \makecell{Train\\Time} \\
    \midrule
      mT5 base  &  264M & 8 & 6e-4 & ~~4h~~ \\
      umT5 base &  269M & 8 & 6e-4 & ~~4h~~ \\
      mT5 large &  538M & 8 & 6e-4 & ~~\hphantom{.}9.5h\\
      mT5 xl    & 1593M & 6 & 5e-4 & \hphantom{.}22.5h\\
      umT5 xl   & 1605M & 6 & 5e-4 & \hphantom{.}22.5h\\
      mT5 xxl   & 5393M & 6 & 5e-4 & 33h~~ \\
      umT5 xxl  & 5417M & 6 & 5e-4 & 33h~~ \\
    \bottomrule
  \end{tabular*}
  \caption{Properties of mT5 encoder models used. The training time is
  measured for 15 epochs 10k updates each using a single A100 GPU, with the
  exception of the xxl models, which are trained using a single H100 GPU.}
  \label{tab:mt5_model_variants}
\end{table}

\looseness-1
\textbf{Training}~
For the shared task submission, we train 13 multilingual models based on
umT5-xl~\citep{chung-etal-2023-unimax}, differing only in random
initialization and whether we express corpus size during sampling using sentences or words.
The sentences are sampled proportionally to the square root of the corpus
size; for ablations, we consider also values of this \textit{sampling
ratio} different from 0.5.

Every model is trained for 15 epochs with 10k batches each, with every
batch consisting of 6 sentences. The model is trained using the
AdaFactor optimizer~\citep{shazeer-etal-2018-adafactor}. The learning rate
follows a warmup schedule: linear increase to 5e-4 during the initial
10\% of training, followed by a cosine decay~\citep{loshchilov-etal-2017-sgdr}
to 0. The model trains for 22.5 hours on a single A100 GPU with 40GB RAM.
For ablation experiments, we also consider other umT5 and
mT5~\citep{xue-etal-2021-mt5} models, whose properties and corresponding
hyperparameters are summarized in Table~\ref{tab:mt5_model_variants}.

For each model, we save checkpoints after every epoch, obtaining a pool of
$13\cdot15$ checkpoints.

\begin{table}[t]
  \input{table_official_metrics.tex}
  \caption{Official results of \CRAC~on the minitest set with various metrics in \%.}
  \label{tab:official_metrics}
\end{table}

\section{Shared Task Results}

In the shared task, teams were permitted to submit up to three systems. We
selected the following configurations based on our checkpoint selection
strategy:
\begin{citemize}
  \item \textbf{CorPipeSingle}, a single best-performing checkpoint selected
    based on overall minidev performance across all corpora;
  \item \textbf{CorPipeBestDev}, employing corpus-specific optimal checkpoints
    selected individually based on minidev performance for each corpus from the pool
    of $13\cdot15$ checkpoints;
  \item \textbf{CorPipeEnsemble}, an ensemble of 5 best-performing checkpoints
    based on overall minidev performance across all corpora.
\end{citemize}
\noindent The first configuration CorPipeSingle corresponds to
practical deployment, where a single model handles all corpora, while the others
aim at maximizing performance.

\begin{table*}[t!]
  \input{table_official_treebanks.tex}
  \caption{Official results of \CRAC~on the minitest set (CoNLL score in \%). The systems $^\dagger$ are described in \citet{prazak-etal-2021-multilingual}; the rest in \citet{novak-etal-2025-findings}.}
  \label{tab:official_treebanks}
\end{table*}

\begin{table*}[t!]
  \input{table_test_ablations.tex}
  \caption{Additional experiments on the CorefUD 1.3 minitest set (CoNLL score in \%). The models in italics are post-competition submissions (i.e., submitted after the shared task deadline).}
  \label{tab:test_ablations}
\end{table*}

The official results of the \CRAC are summarized in
Table~\ref{tab:official_treebanks} showing the CoNLL score
and individual corpora performance, and in Table~\ref{tab:official_metrics}
showing four metrics across all corpora. All CorPipe 25 configurations
substantially surpass all other participants, by 7 percent points for
CorPipeSingle and 8 for CorPipeEnsemble. The CorPipeBestDev configuration
only marginally outperforms CorPipeSingle, which we attribute to the exclusion
of the two smallest corpora this year.

We evaluate additional mT5 and umT5 models on the minitest in
Table~\ref{tab:test_ablations}. The xxl-sized models provide a boost
of more than 1 percent point over the xl size; the ensemble of 3 mT5-xxl and
umT5-xxl models provide an additional 1 percent point gain, achieving the best performance of 77.2\%, a~1.4 percent point
increase compared to the best competition submission.

\begin{table*}[t!]
  \input{table_dev_models.tex}
  \caption{Ablations experiments on the CorefUD 1.3 minidev set (CoNLL score in \%). The results are averages of 3 or more runs and for every run the epoch with best average score over the whole CorefUD is used.}
  \label{tab:dev_models}
\end{table*}

\begin{table*}[t!]
  \input{table_dev_ablations.tex}
  \caption{Ablations experiments on the CorefUD 1.3 minidev set (CoNLL score in \%). The results are averages of 3 or more runs and for every run the epoch with best average score over the whole CorefUD is used.}
  \label{tab:dev_ablations}
\end{table*}

\begin{table*}[t]
  \setlength{\tabcolsep}{4.8pt}
  \begin{tabular*}{\hsize}{l|ccc|ccccc}
    \toprule
      \multirow{3}{*}{Model} & \multicolumn{3}{c|}{TensorFlow} & \multicolumn{5}{c}{PyTorch} \\
                             & Compile & Training & Max & Cold-start & Warm-start & Eager & Compiled & Max \\
                             & time & throughput & batch & compile & compile & throughput & throughput & batch \\
    \midrule
      mT5 base  & 50s & 7.1batch/s & 39 & 55s & 27s & 8.0batch/s & \llap{1}0.3batch/s & 58 \\
      mT5 large & 91s & 3.1batch/s & 13 & 92s & 50s & 3.3batch/s & 4.4batch/s         & 21 \\
      mT5 xl    & 95s & 2.5batch/s &~~5 & 97s & 51s & 2.3batch/s & 2.9batch/s         &~~9 \\
    \bottomrule
  \end{tabular*}
  \caption{Comparison of compilation and training times of CorPipe using the
  latest TensorFlow 2.19 and PyTorch 2.7 with the latest transformers 4.52.4
  on a single A100 40GB GPU. The training throughput is measured using batch
  size of 4 for the xl model and 8 otherwise.}
  \label{tab:tf_pytorch_comparison}
\end{table*}

\section{Ablations Experiments}

We perform a series of ablation experiments on the CorefUD 1.3 minidev set
(to avoid overfitting on the minitest set). The presented results are averages of 3 or more
runs, and for every run the epoch with the best average score across all
corpora is used.

For reference, the minidev scores of the systems submitted to the \CRAC
are summarized in Table~\ref{tab:dev_models}.A.

The first set of experiments evaluates the impact of different models
beyond the mT5 and umT5 families. Notably, we also evaluate the
XLM-RoBERTa-base and XLM-RoBERTa-large
models~\citep{conneau-etal-2020-unsupervised}, the RemBERT
model~\citep{chung-etal-2021-rethinking},
InfoXLM-large~\citep{chi-etal-2021-infoxlm}, and several variants
of the recently introduced T5Gemma
model~\citep{zhang-etal-2025-encoder-decoder}.

\looseness-1
The results are summarized in Table~\ref{tab:dev_models}.B. The umT5 models
consistently outperform the mT5 ones, which is why we used them
in the official submission.\footnote{In this context, it is unfortunate that
the umT5-large model has not been released as it would likely outperform the
mT5-large model, which is a size very suitable for deployment.}
The mT5 and umT5 models outperform the other evaluated models, particularly
because they support longer contexts (Table~\ref{tab:dev_ablations}.C and \citealp{straka-2023-ufal}, Table 4).
When restricting the context to 512 tokens, XLM-RoBERTa-large model
achieves the best performance, surpassing both InfoXLM-large and RemBERT.
Finally, the recently introduced T5Gemma encoder-decoder model adapted from the
Gemma decoder-only model seems to lag behind the umT5 models of corresponding
sizes, despite supporting longer contexts too.

\textbf{Cross-Lingual Zero-Shot Evaluation}~~
Given that our model is multilingual, it can be used to perform coreference
resolution in languages not exposed to during training. In order to evaluate the
performance of our model in such a setting, we train several multilingual
models on corpora from all but one language, and then evaluate their performance
on the excluded corpora. The results are summarized in
Table~\ref{tab:dev_ablations}.A for the mT5-large model and in
Table~\ref{tab:dev_ablations}.B for the umT5-xl model. While the cross-lingual
zero-shot performance is substantially lower by roughly 14 percentage points,
it is still higher than the baseline system
of~\citet{prazak-etal-2021-multilingual} and on par with the best LLM-track
submission. Interestingly, the performance of umT5-xl is higher by more than
2 points, an increase consistent with the results in the supervised
setting.

\textbf{Segment Size}~~
The effect of context larger than the usual 512 tokens is quantified
in Table~\ref{tab:dev_ablations}.C for the mT5-large model and in
Table~\ref{tab:dev_ablations}.D for the umT5-xl model. The results show
that the increase from 512 to 1\,024 tokens leads to a significant performance
increase of more than 2 percentage points, and the further increase to 2\,560
tokens brings a smaller increase by less than 0.5 points.

\looseness-1
\textbf{Sampling Ratio}~~
During training, we sample sentences from the training corpora proportionally
to the square root of their size, following for example
\citet{van-der-goot-etal-2021-massive,straka-2024-corpipe,straka-etal-2024-ufal}.
We quantify the impact of using different exponents (sampling ratios) in
Table~\ref{tab:dev_ablations}.E for the mT5-large model and in
Table~\ref{tab:dev_ablations}.F for the umT5-xl model. The results show that
while the choice of 0.5 is reasonable, the sampling ratio has very little
impact on the average performance. However, we can see a minor effect of the
sampling ratio on the performance of the two largest corpora (the Czech ones),
with the decrease of 0.5 to 1.5 percentage points for uniform sampling (sampling
ratio 0) to the increase of 0.3 to 0.5 percentage points for proportional
sampling (sampling ratio 1).

\section{PyTorch vs TensorFlow}

Having both PyTorch and TensorFlow implementations of CorPipe, we can compare
the two variants in terms of training throughput and memory usage. To this
end, we compare the CorPipe 23 using the latest TensorFlow 2.19 and CorPipe 25
utilizing the latest PyTorch 2.7, both with the latest transformers library
4.52.4, on a single A100 40GB GPU.

The results are presented in Table~\ref{tab:tf_pytorch_comparison}. For all
the base, large, and xl sizes, the PyTorch implementation outperforms the
TensorFlow implementation:
\begin{citemize}
  \item The training throughput is higher by 16\% for the xxl model up to
    45\% for the base model, when comparing compiled PyTorch models
    to compiled TensorFlow models.
  \item The PyTorch model cold-start compilation time is quite similar
    to TensorFlow; however, the warm-start compilation (reusing cached
    compilation files from preceding executions; happens automatically) is significantly
    shorter, being circa half of the TensorFlow time.
  \item The eager PyTorch model has comparable or slightly better performance
    than the compiled TensorFlow model.
  \item The PyTorch implementation has lower memory requirements, allowing
    batches larger by at least 50\% to fit into the GPU memory.
\end{citemize}
\noindent Note that the difference might stem just from different mT5
implementations (FlashAttention, etc.), not necessarily from the frameworks
themselves.

\section{Conclusions}

We introduced CorPipe 25, the winning submission to the \CRAClong~\citep{novak-etal-2025-findings}.
Our approach employs a three-stage pipeline architecture that first predicts empty nodes using
a dedicated pretrained encoder model, then performs mention detection and coreference linking
through a jointly trained system utilizing another pretrained encoder. This complete PyTorch
reimplementation significantly outperforms all other submissions by substantial margins of 7 and 8
percentage points for our single model and ensemble variants, respectively.
The source code and trained models are publicly available at
\hbox{\small\url{https://github.com/ufal/crac2025-corpipe}}.

\section*{Acknowledgements}

Our research has been supported by the OP~JAK project
CZ.02.01.01/00/23\_020/0008518 of the Ministry of Education, Youth and Sports
of the Czech Republic and uses data provided by the \hbox{LINDAT/CLARIAH-CZ}
\hbox{Research} Infrastructure (https://lindat.cz), supported by the Ministry
of Education, Youth and Sports of the Czech Republic (Project No. LM2023062).

\bibliography{anthology,custom}

\end{document}

%% file: table_official_metrics.tex
  \footnotesize\setlength{\tabcolsep}{3pt}
  \renewcommand\cellset{\renewcommand\arraystretch{0.85}}
  \catcode`@ = 13\def@{\bfseries}
  \begin{tabular*}{\hsize}{l*{4}{@{\extracolsep{\fill}}c}}
    \toprule
      System & \makecell{Head-\\match} & \makecell{Partial-\\match} & \makecell{Exact-\\match} & \makecell{With Sin-\\gletons} \\
  \midrule
  \multicolumn{5}{l}{\textsc{Unconstrained}} \\[2pt]
    ~~@CorPipeEnsemble &@\makecell[c]{75.84 \\ 1} &@\makecell[c]{74.90 \\ 1} &@\makecell[c]{72.76 \\ 1} &@\makecell[c]{78.33 \\ 1} \\[7pt]
    ~~@CorPipeBestDev  & \makecell[c]{75.06 \\ 2} & \makecell[c]{74.08 \\ 2} & \makecell[c]{71.97 \\ 2} & \makecell[c]{77.63 \\ 2} \\[7pt]
    ~~@CorPipeSingle   & \makecell[c]{74.75 \\ 3} & \makecell[c]{73.74 \\ 3} & \makecell[c]{71.53 \\ 3} & \makecell[c]{77.43 \\ 3} \\[7pt]
    ~~Stanza           & \makecell[c]{67.81 \\ 4} & \makecell[c]{67.03 \\ 4} & \makecell[c]{64.68 \\ 4} & \makecell[c]{70.64 \\ 4} \\[7pt]
    ~~GLaRef-Propp     & \makecell[c]{61.57 \\ 5} & \makecell[c]{60.72 \\ 5} & \makecell[c]{58.43 \\ 5} & \makecell[c]{65.28 \\ 5} \\[7pt]
    ~~BASELINE-GZ      & \makecell[c]{58.18 \\ 6} & \makecell[c]{57.75 \\ 6} & \makecell[c]{56.48 \\ 6} & \makecell[c]{49.88 \\ 6} \\[7pt]
    ~~BASELINE         & \makecell[c]{56.01 \\ 7} & \makecell[c]{55.58 \\ 7} & \makecell[c]{54.24 \\ 7} & \makecell[c]{47.88 \\ 7} \\

  \midrule
  \multicolumn{5}{l}{\textsc{LLM}} \\[2pt]
    ~~GLaRef-CRAC25    & \makecell[c]{62.96 \\ 1} & \makecell[c]{61.66 \\ 1} & \makecell[c]{58.98 \\ 1} & \makecell[c]{65.61 \\ 1} \\[7pt]
    ~~NUST-FewShot     & \makecell[c]{61.74 \\ 2} & \makecell[c]{61.14 \\ 2} & \makecell[c]{56.34 \\ 2} & \makecell[c]{63.44 \\ 2} \\[7pt]
    ~~PUXCRAC2025      & \makecell[c]{60.09 \\ 3} & \makecell[c]{59.68 \\ 3} & \makecell[c]{55.22 \\ 3} & \makecell[c]{54.77 \\ 4} \\[7pt]
    ~~UWB              & \makecell[c]{59.84 \\ 4} & \makecell[c]{59.55 \\ 4} & \makecell[c]{38.81 \\ 4} & \makecell[c]{62.77 \\ 3} \\
  \bottomrule
\end{tabular*}

%% file: table_official_treebanks.tex
  \scriptsize\setlength{\tabcolsep}{3pt}
  \renewcommand\cellset{\renewcommand\arraystretch{0.70}}
  \catcode`@ = 13\def@{\bfseries}
  \begin{tabular*}{\hsize}{l*{232}{@{\extracolsep{\fill}}c}}
    \toprule
      System & Avg &
      \texttt{ca} &
      \makecell[c]{\texttt{cs} \\ \texttt{\kern-.1em pced\kern-.1em}} &
      \makecell[c]{\texttt{cs} \\ \texttt{pdt}} &
      \texttt{cu} &
      \makecell[c]{\texttt{de} \\ \texttt{\kern-.1em pots\kern-.1em}} &
      \makecell[c]{\texttt{en} \\ \texttt{gum}} &
      \makecell[c]{\texttt{en} \\ \texttt{\kern-.1em litb\kern-.1em}} &
      \texttt{es} &
      \makecell[c]{\texttt{fr} \\ \texttt{\kern-.1em anco\kern-.1em}} &
      \makecell[c]{\texttt{fr} \\ \texttt{\kern-.1em demo\kern-.1em}} &
      \texttt{grc} &
      \texttt{hbo} &
      \makecell[c]{\texttt{hi} \\ \texttt{\kern-.1em hdtb\kern-.1em}} &
      \makecell[c]{\texttt{hu} \\ \texttt{\kern-.1em kork\kern-.1em}} &
      \makecell[c]{\texttt{hu} \\ \texttt{\kern-.1em szeg\kern-.1em}} &
      \texttt{ko} &
      \texttt{lt} &
      \makecell[c]{\texttt{no} \\ \texttt{\kern-.1em bokm\kern-.1em}} &
      \makecell[c]{\texttt{no} \\ \texttt{\kern-.1em nyno\kern-.1em}} &
      \texttt{pl} &
      \texttt{ru} &
      \texttt{tr} \\
  \midrule
  \multicolumn{5}{l}{\textsc{Unconstrained}} \\[2pt]
    ~~@CorPipeEnsemble      &@\makecell[c]{75.8 \\ 1} &@\makecell[c]{82.9 \\ 1} &@\makecell[c]{77.1 \\ 1} &@\makecell[c]{80.7 \\ 1} &@\makecell[c]{65.5 \\ 1} &@\makecell[c]{73.0 \\ 1} &@\makecell[c]{76.1 \\ 1} & \makecell[c]{81.8 \\ 1} &@\makecell[c]{84.5 \\ 1} &@\makecell[c]{76.3 \\ 1} &@\makecell[c]{71.8 \\ 1} &@\makecell[c]{74.5 \\ 1} & \makecell[c]{69.8 \\ 1} &@\makecell[c]{77.7 \\ 1} &@\makecell[c]{68.6 \\ 1} &@\makecell[c]{71.0 \\ 1} &@\makecell[c]{69.9 \\ 1} &@\makecell[c]{77.2 \\ 1} &@\makecell[c]{78.2 \\ 1} &@\makecell[c]{76.3 \\ 1} &@\makecell[c]{80.2 \\ 1} & \makecell[c]{84.2 \\ 3} & \makecell[c]{71.2 \\ 2} \\[6pt]
    ~~@CorPipeBestDev       & \makecell[c]{75.1 \\ 2} & \makecell[c]{82.0 \\ 3} & \makecell[c]{76.3 \\ 2} & \makecell[c]{80.4 \\ 2} & \makecell[c]{62.8 \\ 3} & \makecell[c]{72.6 \\ 3} & \makecell[c]{75.9 \\ 2} & \makecell[c]{81.3 \\ 2} & \makecell[c]{83.8 \\ 3} & \makecell[c]{75.9 \\ 2} & \makecell[c]{69.9 \\ 3} & \makecell[c]{74.3 \\ 3} & \makecell[c]{68.3 \\ 2} & \makecell[c]{77.5 \\ 2} & \makecell[c]{68.3 \\ 2} & \makecell[c]{70.5 \\ 2} & \makecell[c]{69.3 \\ 2} & \makecell[c]{76.0 \\ 2} & \makecell[c]{77.1 \\ 2} & \makecell[c]{74.0 \\ 2} & \makecell[c]{79.9 \\ 2} &@\makecell[c]{84.8 \\ 1} & \makecell[c]{70.4 \\ 3} \\[6pt]
    ~~@CorPipeSingle        & \makecell[c]{74.8 \\ 3} & \makecell[c]{82.5 \\ 2} & \makecell[c]{76.2 \\ 3} & \makecell[c]{80.1 \\ 3} & \makecell[c]{63.0 \\ 2} & \makecell[c]{72.8 \\ 2} & \makecell[c]{75.2 \\ 3} & \makecell[c]{80.8 \\ 3} & \makecell[c]{84.1 \\ 2} & \makecell[c]{75.8 \\ 3} & \makecell[c]{70.3 \\ 2} & \makecell[c]{74.4 \\ 2} & \makecell[c]{66.1 \\ 3} & \makecell[c]{76.5 \\ 3} & \makecell[c]{67.3 \\ 3} & \makecell[c]{69.7 \\ 3} & \makecell[c]{68.9 \\ 3} & \makecell[c]{75.8 \\ 3} & \makecell[c]{76.2 \\ 3} & \makecell[c]{73.6 \\ 3} & \makecell[c]{79.4 \\ 3} & \makecell[c]{84.2 \\ 2} &@\makecell[c]{71.6 \\ 1} \\[6pt]
    ~~Stanza                & \makecell[c]{67.8 \\ 4} & \makecell[c]{79.5 \\ 4} & \makecell[c]{72.7 \\ 4} & \makecell[c]{75.1 \\ 4} & \makecell[c]{40.8 \\ 4} & \makecell[c]{67.3 \\ 4} & \makecell[c]{69.0 \\ 4} & \makecell[c]{74.8 \\ 4} & \makecell[c]{80.4 \\ 4} & \makecell[c]{67.5 \\ 4} & \makecell[c]{62.5 \\ 5} & \makecell[c]{54.9 \\ 4} & \makecell[c]{62.1 \\ 4} & \makecell[c]{74.2 \\ 4} & \makecell[c]{60.0 \\ 4} & \makecell[c]{64.6 \\ 4} & \makecell[c]{67.7 \\ 4} & \makecell[c]{72.8 \\ 4} & \makecell[c]{72.4 \\ 4} & \makecell[c]{71.7 \\ 4} & \makecell[c]{73.0 \\ 4} & \makecell[c]{80.8 \\ 4} & \makecell[c]{47.8 \\ 5} \\[6pt]
    ~~GLaRef-Propp          & \makecell[c]{61.6 \\ 5} & \makecell[c]{68.1 \\ 6} & \makecell[c]{61.7 \\ 6} & \makecell[c]{66.6 \\ 6} & \makecell[c]{39.1 \\ 5} & \makecell[c]{61.2 \\ 5} & \makecell[c]{61.9 \\ 5} & \makecell[c]{70.0 \\ 5} & \makecell[c]{69.1 \\ 7} & \makecell[c]{65.1 \\ 5} & \makecell[c]{66.1 \\ 4} & \makecell[c]{51.3 \\ 5} & \makecell[c]{58.8 \\ 5} & \makecell[c]{69.5 \\ 5} & \makecell[c]{50.9 \\ 5} & \makecell[c]{60.1 \\ 5} & \makecell[c]{60.6 \\ 6} & \makecell[c]{57.6 \\ 7} & \makecell[c]{67.1 \\ 5} & \makecell[c]{66.3 \\ 5} & \makecell[c]{68.0 \\ 6} & \makecell[c]{71.5 \\ 5} & \makecell[c]{44.3 \\ 7} \\[6pt]
    ~~BASELINE-GZ$^\dagger$ & \makecell[c]{58.2 \\ 6} & \makecell[c]{68.8 \\ 5} & \makecell[c]{69.5 \\ 5} & \makecell[c]{67.9 \\ 5} & \makecell[c]{29.5 \\ 6} & \makecell[c]{55.7 \\ 6} & \makecell[c]{61.6 \\ 7} & \makecell[c]{66.0 \\ 6} & \makecell[c]{71.0 \\ 5} & \makecell[c]{63.8 \\ 6} & \makecell[c]{55.0 \\ 6} & \makecell[c]{29.4 \\ 6} & \makecell[c]{31.0 \\ 6} & \makecell[c]{66.8 \\ 6} & \makecell[c]{47.1 \\ 6} & \makecell[c]{54.3 \\ 7} & \makecell[c]{64.3 \\ 5} & \makecell[c]{65.3 \\ 5} & \makecell[c]{62.5 \\ 6} & \makecell[c]{63.0 \\ 6} & \makecell[c]{68.1 \\ 5} & \makecell[c]{67.6 \\ 6} & \makecell[c]{51.7 \\ 4} \\[6pt]
    ~~BASELINE$^\dagger$    & \makecell[c]{56.0 \\ 7} & \makecell[c]{68.0 \\ 7} & \makecell[c]{56.9 \\ 7} & \makecell[c]{63.0 \\ 7} & \makecell[c]{26.3 \\ 7} & \makecell[c]{55.7 \\ 6} & \makecell[c]{61.7 \\ 6} & \makecell[c]{66.0 \\ 6} & \makecell[c]{70.5 \\ 6} & \makecell[c]{63.8 \\ 6} & \makecell[c]{55.0 \\ 6} & \makecell[c]{28.5 \\ 7} & \makecell[c]{31.0 \\ 6} & \makecell[c]{66.8 \\ 6} & \makecell[c]{43.2 \\ 7} & \makecell[c]{54.5 \\ 6} & \makecell[c]{50.3 \\ 7} & \makecell[c]{65.3 \\ 5} & \makecell[c]{62.5 \\ 6} & \makecell[c]{63.0 \\ 6} & \makecell[c]{66.5 \\ 7} & \makecell[c]{67.6 \\ 6} & \makecell[c]{45.9 \\ 6} \\[6pt]

    \noalign{\kern-.45em}\midrule\noalign{\kern-.1em}
  \multicolumn{5}{l}{\textsc{LLM}} \\[2pt]
    ~~GLaRef-CRAC25         & \makecell[c]{63.0 \\ 1} & \makecell[c]{73.5 \\ 2} & \makecell[c]{65.1 \\ 1} & \makecell[c]{71.3 \\ 1} & \makecell[c]{58.2 \\ 2} & \makecell[c]{59.6 \\ 2} & \makecell[c]{58.7 \\ 4} & \makecell[c]{69.0 \\ 4} & \makecell[c]{74.4 \\ 1} & \makecell[c]{66.7 \\ 2} & \makecell[c]{60.4 \\ 2} & \makecell[c]{65.8 \\ 1} & \makecell[c]{44.0 \\ 3} & \makecell[c]{56.4 \\ 4} & \makecell[c]{52.5 \\ 1} & \makecell[c]{59.8 \\ 3} & \makecell[c]{63.0 \\ 3} & \makecell[c]{62.5 \\ 3} & \makecell[c]{64.7 \\ 4} & \makecell[c]{61.6 \\ 4} & \makecell[c]{72.5 \\ 1} & \makecell[c]{68.8 \\ 3} & \makecell[c]{56.2 \\ 2} \\[6pt]
    ~~NUST-FewShot          & \makecell[c]{61.7 \\ 2} & \makecell[c]{60.9 \\ 4} & \makecell[c]{51.4 \\ 4} & \makecell[c]{54.3 \\ 4} & \makecell[c]{58.5 \\ 1} & \makecell[c]{48.7 \\ 4} & \makecell[c]{69.8 \\ 2} & \makecell[c]{70.4 \\ 2} & \makecell[c]{61.8 \\ 4} & \makecell[c]{71.9 \\ 1} & \makecell[c]{57.6 \\ 3} & \makecell[c]{57.9 \\ 2} &@\makecell[c]{80.2 \\ 1} & \makecell[c]{71.3 \\ 2} & \makecell[c]{43.5 \\ 3} & \makecell[c]{52.3 \\ 4} & \makecell[c]{66.0 \\ 2} & \makecell[c]{59.2 \\ 4} & \makecell[c]{72.8 \\ 2} & \makecell[c]{68.9 \\ 2} & \makecell[c]{70.8 \\ 2} & \makecell[c]{71.4 \\ 2} & \makecell[c]{39.0 \\ 3} \\[6pt]
    ~~PUXCRAC2025           & \makecell[c]{60.1 \\ 3} & \makecell[c]{68.0 \\ 3} & \makecell[c]{56.9 \\ 3} & \makecell[c]{63.0 \\ 3} & \makecell[c]{43.7 \\ 3} & \makecell[c]{57.4 \\ 3} & \makecell[c]{61.7 \\ 3} & \makecell[c]{69.1 \\ 3} & \makecell[c]{70.5 \\ 3} & \makecell[c]{63.8 \\ 3} & \makecell[c]{61.5 \\ 1} & \makecell[c]{47.9 \\ 3} & \makecell[c]{45.3 \\ 2} & \makecell[c]{66.8 \\ 3} & \makecell[c]{50.6 \\ 2} & \makecell[c]{61.6 \\ 2} & \makecell[c]{50.3 \\ 4} & \makecell[c]{65.3 \\ 1} & \makecell[c]{65.2 \\ 3} & \makecell[c]{63.0 \\ 3} & \makecell[c]{66.5 \\ 3} & \makecell[c]{67.6 \\ 4} & \makecell[c]{56.1 \\ 1} \\[6pt]
    ~~UWB                   & \makecell[c]{59.8 \\ 4} & \makecell[c]{79.2 \\ 1} & \makecell[c]{61.0 \\ 2} & \makecell[c]{68.2 \\ 2} & \makecell[c]{25.3 \\ 4} & \makecell[c]{67.6 \\ 1} & \makecell[c]{73.6 \\ 1} &@\makecell[c]{84.0 \\ 1} & \makecell[c]{73.6 \\ 2} & \makecell[c]{58.6 \\ 4} & \makecell[c]{49.1 \\ 4} & \makecell[c]{47.6 \\ 4} & \makecell[c]{~~0.0\\ 4} & \makecell[c]{75.8 \\ 1} & \makecell[c]{38.9 \\ 4} & \makecell[c]{67.3 \\ 1} & \makecell[c]{68.3 \\ 1} & \makecell[c]{63.4 \\ 2} & \makecell[c]{73.8 \\ 1} & \makecell[c]{72.0 \\ 1} & \makecell[c]{64.5 \\ 4} & \makecell[c]{80.1 \\ 1} & \makecell[c]{24.3 \\ 4} \\[6pt]

    \noalign{\kern-.55em}\bottomrule
\end{tabular*}

%% file: table_test_ablations.tex
    \scriptsize\setlength{\tabcolsep}{1pt}
    \catcode`@ = 13\def@{\bfseries}
    \catcode`! = 13\def!{\itshape}
    \begin{tabular*}{\hsize}{l*{23}{@{\extracolsep{\fill}}c}}
      \toprule
        System & Avg &
        \texttt{ca} &
        \makecell[c]{\texttt{cs} \\ \texttt{\kern-.1em pced\kern-.1em}} &
        \makecell[c]{\texttt{cs} \\ \texttt{pdt}} &
        \texttt{cu} &
        \makecell[c]{\texttt{de} \\ \texttt{\kern-.1em pots\kern-.1em}} &
        \makecell[c]{\texttt{en} \\ \texttt{gum}} &
        \makecell[c]{\texttt{en} \\ \texttt{\kern-.1em litb\kern-.1em}} &
        \texttt{es} &
        \makecell[c]{\texttt{fr} \\ \texttt{\kern-.1em anco\kern-.1em}} &
        \makecell[c]{\texttt{fr} \\ \texttt{\kern-.1em demo\kern-.1em}} &
        \texttt{grc} &
        \texttt{hbo} &
        \makecell[c]{\texttt{hi} \\ \texttt{\kern-.1em hdtb\kern-.1em}} &
        \makecell[c]{\texttt{hu} \\ \texttt{\kern-.1em kork\kern-.1em}} &
        \makecell[c]{\texttt{hu} \\ \texttt{\kern-.1em szeg\kern-.1em}} &
        \texttt{ko} &
        \texttt{lt} &
        \makecell[c]{\texttt{no} \\ \texttt{\kern-.1em bokm\kern-.1em}} &
        \makecell[c]{\texttt{no} \\ \texttt{\kern-.1em nyno\kern-.1em}} &
        \texttt{pl} &
        \texttt{ru} &
        \texttt{tr} \\
    \midrule
\multicolumn{22}{l}{\textsc{A) CorPipe Single Models}} \\[2pt]
~~Single mT5-large model & \textcolor{black}{72.84} & \textcolor{black}{80.1} & \textcolor{black}{74.6} & \textcolor{black}{78.0} & \textcolor{black}{58.5} & \textcolor{black}{67.2} & \textcolor{black}{73.3} & \textcolor{black}{77.4} & \textcolor{black}{82.0} & \textcolor{black}{72.1} & \textcolor{black}{68.5} & \textcolor{black}{71.2} & \textcolor{black}{67.9} & \textcolor{black}{76.3} & \textcolor{black}{67.3} & \textcolor{black}{68.0} & \textcolor{black}{69.8} & \textcolor{black}{74.4} & \textcolor{black}{75.2} & \textcolor{black}{74.0} & \textcolor{black}{77.5} & \textcolor{black}{81.2} & \textcolor{black}{67.7} \\[2pt]
~~\multirow{2}{*}{!Single umT5-base model} & \textcolor{red!100.0!black}{--\kern 0.04em 3.54} & \textcolor{red!100.0!black}{--\kern 0.04em 2.7} & \textcolor{red!100.0!black}{--\kern 0.04em 1.1} & \textcolor{red!100.0!black}{--\kern 0.04em 2.9} & \textcolor{red!100.0!black}{--\kern 0.04em 5.0} & \textcolor{red!100.0!black}{--\kern 0.04em 5.2} & \textcolor{red!100.0!black}{--\kern 0.04em 2.3} & \textcolor{red!100.0!black}{--\kern 0.04em 4.6} & \textcolor{red!100.0!black}{--\kern 0.04em 3.5} & \textcolor{red!100.0!black}{--\kern 0.04em 0.9} & \textcolor{red!100.0!black}{--\kern 0.04em 1.8} & \textcolor{red!100.0!black}{--\kern 0.04em 6.3} & \textcolor{red!100.0!black}{--\kern 0.04em 8.9} & \textcolor{red!100.0!black}{--\kern 0.04em 3.6} & \textcolor{red!100.0!black}{--\kern 0.04em 5.8} & \textcolor{red!100.0!black}{--\kern 0.04em 4.3} & \textcolor{red!100.0!black}{--\kern 0.04em 2.0} & \textcolor{red!100.0!black}{--\kern 0.04em 1.5} & \textcolor{red!100.0!black}{--\kern 0.04em 2.0} & \textcolor{red!100.0!black}{--\kern 0.04em 3.5} & \textcolor{red!100.0!black}{--\kern 0.04em 3.0} & \textcolor{red!100.0!black}{--\kern 0.04em 3.5} & \textcolor{red!100.0!black}{--\kern 0.04em 3.8} \\
~~ & \textcolor{black}{69.27} & \textcolor{black}{77.4} & \textcolor{black}{73.5} & \textcolor{black}{75.1} & \textcolor{black}{53.5} & \textcolor{black}{62.0} & \textcolor{black}{71.0} & \textcolor{black}{72.8} & \textcolor{black}{78.6} & \textcolor{black}{71.2} & \textcolor{black}{66.7} & \textcolor{black}{64.9} & \textcolor{black}{59.0} & \textcolor{black}{72.7} & \textcolor{black}{61.5} & \textcolor{black}{63.7} & \textcolor{black}{67.8} & \textcolor{black}{72.9} & \textcolor{black}{73.2} & \textcolor{black}{70.4} & \textcolor{black}{74.5} & \textcolor{black}{77.8} & \textcolor{black}{63.9} \\[2pt]
~~\multirow{2}{*}{Single umT5-xl model} & \textcolor{blue!56.0!black}{+1.96} & \textcolor{blue!78.5!black}{+2.4} & \textcolor{blue!55.7!black}{+1.6} & \textcolor{blue!62.2!black}{+2.1} & \textcolor{blue!41.4!black}{+4.5} & \textcolor{blue!91.8!black}{+5.6} & \textcolor{blue!51.5!black}{+1.9} & \textcolor{blue!50.2!black}{+3.4} & \textcolor{blue!86.0!black}{+2.0} & \textcolor{blue!67.9!black}{+3.7} & \textcolor{blue!47.5!black}{+1.8} & \textcolor{blue!45.8!black}{+3.2} & \textcolor{red!20.8!black}{--\kern 0.04em 1.8} & \textcolor{blue!13.7!black}{+0.2} & \textcolor{blue!0.8!black}{--\kern 0.04em 0.0} & \textcolor{blue!44.5!black}{+1.7} & \textcolor{red!45.6!black}{--\kern 0.04em 0.9} & \textcolor{blue!43.7!black}{+1.4} & \textcolor{blue!33.9!black}{+1.0} & \textcolor{red!9.2!black}{--\kern 0.04em 0.4} & \textcolor{blue!69.0!black}{+1.9} & \textcolor{blue!78.9!black}{+3.0} & \textcolor{blue!83.9!black}{+3.9} \\
~~ & \textcolor{black}{74.75} & \textcolor{black}{82.5} & \textcolor{black}{76.2} & \textcolor{black}{80.1} & \textcolor{black}{63.0} & \textcolor{black}{72.8} & \textcolor{black}{75.2} & \textcolor{black}{80.8} & \textcolor{black}{84.1} & \textcolor{black}{75.8} & \textcolor{black}{70.3} & \textcolor{black}{74.4} & \textcolor{black}{66.1} & \textcolor{black}{76.5} & \textcolor{black}{67.3} & \textcolor{black}{69.7} & \textcolor{black}{68.9} & \textcolor{black}{75.8} & \textcolor{black}{76.2} & \textcolor{black}{73.6} & \textcolor{black}{79.4} & \textcolor{black}{84.2} & \textcolor{black}{71.6} \\[2pt]
~~\multirow{2}{*}{!Single mT5-xxl model} & \textcolor{blue!93.6!black}{+3.16} & \textcolor{blue!100.0!black}{@+3.0} & \textcolor{blue!100.0!black}{@+2.9} & \textcolor{blue!100.0!black}{@+3.3} & \textcolor{blue!100.0!black}{@+10.8} & \textcolor{blue!100.0!black}{@+6.1} & \textcolor{blue!94.4!black}{+3.6} & \textcolor{blue!74.1!black}{+5.0} & \textcolor{blue!95.9!black}{+2.4} & \textcolor{blue!66.4!black}{+3.6} & \textcolor{blue!100.0!black}{@+4.0} & \textcolor{blue!92.2!black}{+6.2} & \textcolor{blue!100.0!black}{@+6.5} & \textcolor{blue!28.2!black}{+0.4} & \textcolor{red!88.2!black}{--\kern 0.04em 5.1} & \textcolor{blue!100.0!black}{@+3.9} & \textcolor{red!25.0!black}{--\kern 0.04em 0.5} & \textcolor{blue!100.0!black}{@+3.2} & \textcolor{blue!56.8!black}{+1.7} & \textcolor{blue!59.7!black}{+1.5} & \textcolor{blue!97.2!black}{+2.7} & \textcolor{blue!97.6!black}{+3.7} & \textcolor{blue!40.5!black}{+1.9} \\
~~ & \textcolor{black}{76.04} & \textcolor{black}{@83.1} & \textcolor{black}{@77.5} & \textcolor{black}{@81.3} & \textcolor{black}{@69.3} & \textcolor{black}{@73.3} & \textcolor{black}{76.9} & \textcolor{black}{82.4} & \textcolor{black}{84.4} & \textcolor{black}{75.7} & \textcolor{black}{@72.4} & \textcolor{black}{77.5} & \textcolor{black}{@74.4} & \textcolor{black}{76.7} & \textcolor{black}{62.2} & \textcolor{black}{@71.9} & \textcolor{black}{69.3} & \textcolor{black}{@77.6} & \textcolor{black}{76.9} & \textcolor{black}{75.4} & \textcolor{black}{80.2} & \textcolor{black}{84.9} & \textcolor{black}{69.6} \\[2pt]
~~\multirow{2}{*}{!Single umT5-xxl model} & \textcolor{blue!100.0!black}{@+3.46} & \textcolor{blue!99.7!black}{+3.0} & \textcolor{blue!94.5!black}{+2.7} & \textcolor{blue!98.5!black}{+3.2} & \textcolor{blue!84.5!black}{+9.1} & \textcolor{blue!73.1!black}{+4.5} & \textcolor{blue!100.0!black}{@+3.8} & \textcolor{blue!100.0!black}{@+6.8} & \textcolor{blue!95.9!black}{+2.4} & \textcolor{blue!100.0!black}{@+5.4} & \textcolor{blue!76.0!black}{+3.0} & \textcolor{blue!100.0!black}{@+6.8} & \textcolor{blue!72.6!black}{+4.8} & \textcolor{blue!99.2!black}{+1.4} & \textcolor{red!20.4!black}{--\kern 0.04em 1.2} & \textcolor{blue!82.4!black}{+3.2} & \textcolor{red!30.9!black}{--\kern 0.04em 0.6} & \textcolor{blue!20.3!black}{+0.6} & \textcolor{blue!100.0!black}{@+3.1} & \textcolor{blue!100.0!black}{@+2.3} & \textcolor{blue!100.0!black}{@+2.8} & \textcolor{blue!100.0!black}{@+3.8} & \textcolor{blue!100.0!black}{@+4.7} \\
~~ & \textcolor{black}{@76.26} & \textcolor{black}{83.1} & \textcolor{black}{77.3} & \textcolor{black}{81.2} & \textcolor{black}{67.6} & \textcolor{black}{71.7} & \textcolor{black}{@77.1} & \textcolor{black}{@84.2} & \textcolor{black}{84.4} & \textcolor{black}{@77.5} & \textcolor{black}{71.5} & \textcolor{black}{@78.1} & \textcolor{black}{72.7} & \textcolor{black}{77.7} & \textcolor{black}{66.1} & \textcolor{black}{71.2} & \textcolor{black}{69.2} & \textcolor{black}{75.0} & \textcolor{black}{@78.3} & \textcolor{black}{@76.3} & \textcolor{black}{@80.3} & \textcolor{black}{@85.0} & \textcolor{black}{@72.4} \\
    \midrule
\multicolumn{22}{l}{\textsc{B) CorPipe Ensemble Models}} \\[2pt]
~~Single umT5-xl model & \textcolor{black}{74.75} & \textcolor{black}{82.5} & \textcolor{black}{76.2} & \textcolor{black}{80.1} & \textcolor{black}{63.0} & \textcolor{black}{72.8} & \textcolor{black}{75.2} & \textcolor{black}{80.8} & \textcolor{black}{84.1} & \textcolor{black}{75.8} & \textcolor{black}{70.3} & \textcolor{black}{74.4} & \textcolor{black}{66.1} & \textcolor{black}{76.5} & \textcolor{black}{67.3} & \textcolor{black}{69.7} & \textcolor{black}{68.9} & \textcolor{black}{75.8} & \textcolor{black}{76.2} & \textcolor{black}{73.6} & \textcolor{black}{79.4} & \textcolor{black}{84.2} & \textcolor{black}{71.6} \\[2pt]
~~\multirow{2}{*}{5 umT5-xl models} & \textcolor{blue!44.3!black}{+1.05} & \textcolor{blue!24.6!black}{+0.4} & \textcolor{blue!49.4!black}{+0.9} & \textcolor{blue!38.4!black}{+0.6} & \textcolor{blue!35.0!black}{+2.5} & \textcolor{blue!8.0!black}{+0.2} & \textcolor{blue!31.5!black}{+0.8} & \textcolor{blue!31.6!black}{+1.0} & \textcolor{blue!26.8!black}{+0.4} & \textcolor{blue!24.9!black}{+0.5} & \textcolor{blue!58.5!black}{+1.5} & \textcolor{blue!2.2!black}{+0.1} & \textcolor{blue!42.0!black}{+3.7} & \textcolor{blue!73.9!black}{+1.2} & \textcolor{blue!100.0!black}{@+1.3} & \textcolor{blue!35.9!black}{+1.3} & \textcolor{blue!61.9!black}{+1.0} & \textcolor{blue!87.0!black}{+1.4} & \textcolor{blue!100.0!black}{@+2.0} & \textcolor{blue!100.0!black}{@+2.7} & \textcolor{blue!38.6!black}{+0.8} & \textcolor{red!100.0!black}{--\kern 0.04em 0.0} & \textcolor{red!64.1!black}{--\kern 0.04em 0.4} \\
~~ & \textcolor{black}{75.84} & \textcolor{black}{82.9} & \textcolor{black}{77.1} & \textcolor{black}{80.7} & \textcolor{black}{65.5} & \textcolor{black}{73.0} & \textcolor{black}{76.1} & \textcolor{black}{81.8} & \textcolor{black}{84.5} & \textcolor{black}{76.3} & \textcolor{black}{71.8} & \textcolor{black}{74.5} & \textcolor{black}{69.8} & \textcolor{black}{77.7} & \textcolor{black}{@68.6} & \textcolor{black}{71.0} & \textcolor{black}{69.9} & \textcolor{black}{77.2} & \textcolor{black}{@78.2} & \textcolor{black}{@76.3} & \textcolor{black}{80.2} & \textcolor{black}{84.2} & \textcolor{black}{71.2} \\[2pt]
~~\multirow{2}{*}{!3 mT5-xxl models} & \textcolor{blue!88.8!black}{+2.15} & \textcolor{blue!81.1!black}{+1.4} & \textcolor{blue!75.0!black}{+1.4} & \textcolor{blue!62.8!black}{+1.0} & \textcolor{blue!97.7!black}{+7.0} & \textcolor{blue!100.0!black}{@+2.1} & \textcolor{blue!79.8!black}{+2.0} & \textcolor{blue!85.8!black}{+2.9} & \textcolor{blue!74.8!black}{+1.0} & \textcolor{blue!39.3!black}{+1.0} & \textcolor{blue!92.5!black}{+2.4} & \textcolor{blue!98.4!black}{+6.1} & \textcolor{blue!100.0!black}{@+8.9} & \textcolor{blue!83.7!black}{+1.3} & \textcolor{red!25.5!black}{--\kern 0.04em 0.1} & \textcolor{blue!77.8!black}{+2.8} & \textcolor{blue!66.1!black}{+1.1} & \textcolor{blue!100.0!black}{@+1.6} & \textcolor{blue!23.2!black}{+0.5} & \textcolor{blue!84.8!black}{+2.3} & \textcolor{blue!68.6!black}{+1.4} & \textcolor{blue!33.5!black}{+0.6} & \textcolor{red!100.0!black}{--\kern 0.04em 0.6} \\
~~ & \textcolor{black}{76.93} & \textcolor{black}{83.9} & \textcolor{black}{77.6} & \textcolor{black}{81.1} & \textcolor{black}{70.0} & \textcolor{black}{@74.9} & \textcolor{black}{77.3} & \textcolor{black}{83.7} & \textcolor{black}{85.1} & \textcolor{black}{76.7} & \textcolor{black}{72.7} & \textcolor{black}{80.5} & \textcolor{black}{@75.0} & \textcolor{black}{77.8} & \textcolor{black}{67.2} & \textcolor{black}{72.5} & \textcolor{black}{70.0} & \textcolor{black}{@77.4} & \textcolor{black}{76.7} & \textcolor{black}{75.9} & \textcolor{black}{80.8} & \textcolor{black}{84.8} & \textcolor{black}{71.0} \\[2pt]
~~\multirow{2}{*}{!3 umT5-xxl models} & \textcolor{blue!83.7!black}{+2.05} & \textcolor{blue!63.4!black}{+1.1} & \textcolor{blue!81.7!black}{+1.5} & \textcolor{blue!100.0!black}{@+1.7} & \textcolor{blue!80.1!black}{+5.8} & \textcolor{blue!48.1!black}{+1.0} & \textcolor{blue!75.1!black}{+2.0} & \textcolor{blue!87.0!black}{+2.9} & \textcolor{blue!67.7!black}{+0.9} & \textcolor{blue!100.0!black}{@+2.2} & \textcolor{blue!97.2!black}{+2.5} & \textcolor{blue!62.3!black}{+3.9} & \textcolor{blue!88.3!black}{+7.9} & \textcolor{blue!86.3!black}{+1.3} & \textcolor{red!100.0!black}{--\kern 0.04em 0.5} & \textcolor{blue!100.0!black}{@+3.5} & \textcolor{blue!73.2!black}{+1.2} & \textcolor{red!100.0!black}{--\kern 0.04em 1.0} & \textcolor{blue!54.1!black}{+1.1} & \textcolor{blue!86.7!black}{+2.3} & \textcolor{blue!90.0!black}{+1.9} & \textcolor{blue!100.0!black}{@+1.8} & \textcolor{blue!100.0!black}{@+0.2} \\
~~ & \textcolor{black}{76.80} & \textcolor{black}{83.6} & \textcolor{black}{77.7} & \textcolor{black}{@81.8} & \textcolor{black}{68.8} & \textcolor{black}{73.8} & \textcolor{black}{77.2} & \textcolor{black}{83.7} & \textcolor{black}{85.0} & \textcolor{black}{@78.0} & \textcolor{black}{72.8} & \textcolor{black}{78.3} & \textcolor{black}{74.0} & \textcolor{black}{77.8} & \textcolor{black}{66.8} & \textcolor{black}{@73.2} & \textcolor{black}{70.1} & \textcolor{black}{74.8} & \textcolor{black}{77.3} & \textcolor{black}{75.9} & \textcolor{black}{81.3} & \textcolor{black}{@86.0} & \textcolor{black}{@71.8} \\[2pt]
~~!3 mT5-xxl models + & \textcolor{blue!100.0!black}{@+2.45} & \textcolor{blue!100.0!black}{@+1.7} & \textcolor{blue!100.0!black}{@+1.8} & \textcolor{blue!90.7!black}{+1.5} & \textcolor{blue!100.0!black}{@+7.2} & \textcolor{blue!31.1!black}{+0.6} & \textcolor{blue!100.0!black}{@+2.5} & \textcolor{blue!100.0!black}{@+3.3} & \textcolor{blue!100.0!black}{@+1.3} & \textcolor{blue!82.5!black}{+1.8} & \textcolor{blue!100.0!black}{@+2.6} & \textcolor{blue!100.0!black}{@+6.2} & \textcolor{blue!99.9!black}{+8.9} & \textcolor{blue!100.0!black}{@+1.6} & \textcolor{blue!35.2!black}{+0.5} & \textcolor{blue!76.1!black}{+2.7} & \textcolor{blue!100.0!black}{@+1.6} & \textcolor{blue!9.3!black}{+0.1} & \textcolor{blue!65.5!black}{+1.3} & \textcolor{blue!94.3!black}{+2.5} & \textcolor{blue!100.0!black}{@+2.1} & \textcolor{blue!91.5!black}{+1.6} & \textcolor{blue!95.2!black}{+0.2} \\
~~!+3 umT5-xxl models & \textcolor{black}{@77.20} & \textcolor{black}{@84.2} & \textcolor{black}{@78.0} & \textcolor{black}{81.6} & \textcolor{black}{@70.2} & \textcolor{black}{73.4} & \textcolor{black}{@77.8} & \textcolor{black}{@84.1} & \textcolor{black}{@85.4} & \textcolor{black}{77.6} & \textcolor{black}{@72.9} & \textcolor{black}{@80.6} & \textcolor{black}{75.0} & \textcolor{black}{@78.1} & \textcolor{black}{67.8} & \textcolor{black}{72.4} & \textcolor{black}{@70.5} & \textcolor{black}{75.9} & \textcolor{black}{77.5} & \textcolor{black}{76.1} & \textcolor{black}{@81.5} & \textcolor{black}{85.8} & \textcolor{black}{71.8} \\
    \midrule
\multicolumn{22}{l}{\textsc{C) CorPipe Per-Corpus Best Models}} \\[2pt]
~~Single umT5-xl model & \textcolor{black}{74.75} & \textcolor{black}{@82.5} & \textcolor{black}{76.2} & \textcolor{black}{80.1} & \textcolor{black}{@63.0} & \textcolor{black}{@72.8} & \textcolor{black}{75.2} & \textcolor{black}{80.8} & \textcolor{black}{@84.1} & \textcolor{black}{75.8} & \textcolor{black}{@70.3} & \textcolor{black}{@74.4} & \textcolor{black}{66.1} & \textcolor{black}{76.5} & \textcolor{black}{67.3} & \textcolor{black}{69.7} & \textcolor{black}{68.9} & \textcolor{black}{75.8} & \textcolor{black}{76.2} & \textcolor{black}{73.6} & \textcolor{black}{79.4} & \textcolor{black}{84.2} & \textcolor{black}{@71.6} \\[2pt]
~~\multirow{2}{*}{Per-corpus best umT5-xl model} & \textcolor{blue!100.0!black}{@+0.35} & \textcolor{red!100.0!black}{--\kern 0.04em 0.5} & \textcolor{blue!100.0!black}{@+0.1} & \textcolor{blue!100.0!black}{@+0.3} & \textcolor{red!100.0!black}{--\kern 0.04em 0.2} & \textcolor{red!100.0!black}{--\kern 0.04em 0.2} & \textcolor{blue!100.0!black}{@+0.7} & \textcolor{blue!100.0!black}{@+0.5} & \textcolor{red!100.0!black}{--\kern 0.04em 0.3} & \textcolor{blue!100.0!black}{@+0.2} & \textcolor{red!100.0!black}{--\kern 0.04em 0.4} & \textcolor{red!100.0!black}{--\kern 0.04em 0.1} & \textcolor{blue!100.0!black}{@+2.2} & \textcolor{blue!100.0!black}{@+1.0} & \textcolor{blue!100.0!black}{@+1.0} & \textcolor{blue!100.0!black}{@+0.8} & \textcolor{blue!100.0!black}{@+0.4} & \textcolor{blue!100.0!black}{@+0.2} & \textcolor{blue!100.0!black}{@+0.9} & \textcolor{blue!100.0!black}{@+0.4} & \textcolor{blue!100.0!black}{@+0.5} & \textcolor{blue!100.0!black}{@+0.6} & \textcolor{red!100.0!black}{--\kern 0.04em 1.2} \\
~~ & \textcolor{black}{@75.06} & \textcolor{black}{82.0} & \textcolor{black}{@76.3} & \textcolor{black}{@80.4} & \textcolor{black}{62.8} & \textcolor{black}{72.6} & \textcolor{black}{@75.9} & \textcolor{black}{@81.3} & \textcolor{black}{83.8} & \textcolor{black}{@75.9} & \textcolor{black}{69.9} & \textcolor{black}{74.3} & \textcolor{black}{@68.3} & \textcolor{black}{@77.5} & \textcolor{black}{@68.3} & \textcolor{black}{@70.5} & \textcolor{black}{@69.3} & \textcolor{black}{@76.0} & \textcolor{black}{@77.1} & \textcolor{black}{@74.0} & \textcolor{black}{@79.9} & \textcolor{black}{@84.8} & \textcolor{black}{70.4} \\
    \bottomrule
    \end{tabular*}

%% file: table_dev_models.tex
    \scriptsize\setlength{\tabcolsep}{1pt}
    \catcode`@ = 13\def@{\bfseries}
    \catcode`! = 13\def!{\itshape}
\begin{tabular*}{\hsize}{l*{23}{@{\extracolsep{\fill}}c}}
      \toprule
        System & Avg &
        \texttt{ca} &
        \makecell[c]{\texttt{cs} \\ \texttt{\kern-.1em pced\kern-.1em}} &
        \makecell[c]{\texttt{cs} \\ \texttt{pdt}} &
        \texttt{cu} &
        \makecell[c]{\texttt{de} \\ \texttt{\kern-.1em pots\kern-.1em}} &
        \makecell[c]{\texttt{en} \\ \texttt{gum}} &
        \makecell[c]{\texttt{en} \\ \texttt{\kern-.1em litb\kern-.1em}} &
        \texttt{es} &
        \makecell[c]{\texttt{fr} \\ \texttt{\kern-.1em anco\kern-.1em}} &
        \makecell[c]{\texttt{fr} \\ \texttt{\kern-.1em demo\kern-.1em}} &
        \texttt{grc} &
        \texttt{hbo} &
        \makecell[c]{\texttt{hi} \\ \texttt{\kern-.1em hdtb\kern-.1em}} &
        \makecell[c]{\texttt{hu} \\ \texttt{\kern-.1em kork\kern-.1em}} &
        \makecell[c]{\texttt{hu} \\ \texttt{\kern-.1em szeg\kern-.1em}} &
        \texttt{ko} &
        \texttt{lt} &
        \makecell[c]{\texttt{no} \\ \texttt{\kern-.1em bokm\kern-.1em}} &
        \makecell[c]{\texttt{no} \\ \texttt{\kern-.1em nyno\kern-.1em}} &
        \texttt{pl} &
        \texttt{ru} &
        \texttt{tr} \\
    \midrule
\multicolumn{22}{l}{\textsc{A) Submitted CRAC25 Systems}} \\[2pt]
~~CorPipeEnsemble & \textcolor{black}{@76.51} & \textcolor{black}{@84.1} & \textcolor{black}{@76.9} & \textcolor{black}{@81.1} & \textcolor{black}{@64.2} & \textcolor{black}{77.9} & \textcolor{black}{@77.5} & \textcolor{black}{80.0} & \textcolor{black}{@85.1} & \textcolor{black}{@79.6} & \textcolor{black}{@72.5} & \textcolor{black}{76.1} & \textcolor{black}{@66.8} & \textcolor{black}{@82.0} & \textcolor{black}{@69.7} & \textcolor{black}{@73.1} & \textcolor{black}{@69.4} & \textcolor{black}{@81.6} & \textcolor{black}{@80.1} & \textcolor{black}{79.7} & \textcolor{black}{@80.3} & \textcolor{black}{@80.0} & \textcolor{black}{65.5} \\
\bottomrule
    \end{tabular*}

%% file: table_dev_ablations.tex
    \scriptsize\setlength{\tabcolsep}{1pt}
    \catcode`@ = 13\def@{\bfseries}
    \catcode`! = 13\def!{\itshape}
    \begin{tabular*}{\hsize}{l*{23}{@{\extracolsep{\fill}}c}}
      \toprule
        System & Avg &
        \texttt{ca} &
        \makecell[c]{\texttt{cs} \\ \texttt{\kern-.1em pced\kern-.1em}} &
        \makecell[c]{\texttt{cs} \\ \texttt{pdt}} &
        \texttt{cu} &
        \makecell[c]{\texttt{de} \\ \texttt{\kern-.1em pots\kern-.1em}} &
        \makecell[c]{\texttt{en} \\ \texttt{gum}} &
        \makecell[c]{\texttt{en} \\ \texttt{\kern-.1em litb\kern-.1em}} &
        \texttt{es} &
        \makecell[c]{\texttt{fr} \\ \texttt{\kern-.1em anco\kern-.1em}} &
        \makecell[c]{\texttt{fr} \\ \texttt{\kern-.1em demo\kern-.1em}} &
        \texttt{grc} &
        \texttt{hbo} &
        \makecell[c]{\texttt{hi} \\ \texttt{\kern-.1em hdtb\kern-.1em}} &
        \makecell[c]{\texttt{hu} \\ \texttt{\kern-.1em kork\kern-.1em}} &
        \makecell[c]{\texttt{hu} \\ \texttt{\kern-.1em szeg\kern-.1em}} &
        \texttt{ko} &
        \texttt{lt} &
        \makecell[c]{\texttt{no} \\ \texttt{\kern-.1em bokm\kern-.1em}} &
        \makecell[c]{\texttt{no} \\ \texttt{\kern-.1em nyno\kern-.1em}} &
        \texttt{pl} &
        \texttt{ru} &
        \texttt{tr} \\
    \midrule
\multicolumn{22}{l}{\textsc{A) Cross-Lingual Zero-Shot Evaluation of mT5-large Model}} \\[2pt]
~~Single mT5-large Model & \textcolor{black}{@73.26} & \textcolor{black}{@81.3} & \textcolor{black}{@73.8} & \textcolor{black}{@77.0} & \textcolor{black}{@57.7} & \textcolor{black}{@75.3} & \textcolor{black}{@74.1} & \textcolor{black}{@75.9} & \textcolor{black}{@81.7} & \textcolor{black}{@74.9} & \textcolor{black}{@69.7} & \textcolor{black}{@72.1} & \textcolor{black}{@65.2} & \textcolor{black}{@79.7} & \textcolor{black}{@66.4} & \textcolor{black}{@68.7} & \textcolor{black}{@67.7} & \textcolor{black}{@80.0} & \textcolor{black}{@77.2} & \textcolor{black}{@77.5} & \textcolor{black}{@76.8} & \textcolor{black}{@76.2} & \textcolor{black}{@62.8} \\
\bottomrule
    \end{tabular*}